# An Extensive Repot on Cellular Automata Based Artificial Immune System for Strengthening Automated Protein Prediction

Pokkuluri Kiran Sree[*1], Inampudi Ramesh Babuhor[2], SSSN Usha Devi N[3]

[1] Research Scholar, Dept of CSE, JNTU Hyderabad, India

[2] Professor, Dept of CSE, Acharya Nagarjuna University, Guntur, India

[3] Assistant Professor, Dept of CSE, JNTU Kakinada, India

profkiransree@gmail.com, drirameshbabu@gmail.com, usha.jntuk@gmail.com

*Abstract*

Artificial Immune System (AIS-MACA) a novel computational intelligence technique is can be used for strengthening the automated protein prediction system with more adaptability and incorporating more parallelism to the system. Most of the existing approaches are sequential which will classify the input into four major classes and these are designed for similar sequences. AIS-MACA is designed to identify ten classes from the sequences that share twilight zone similarity and identity with the training sequences with mixed and hybrid variations. This method also predicts three states (helix, strand, and coil) for the secondary structure. Our comprehensive design considers 10 feature selection methods and 4 classifiers to develop MACA (Multiple Attractor Cellular Automata) based classifiers that are build for each of the ten classes. We have tested the proposed classifier with twilight-zone and 1-high-similarity benchmark datasets with over three dozens of modern competing predictors shows that AIS-MACA provides the best overall accuracy that ranges between 80% and 89.8% depending on the dataset.

*Keywords*

*Protein Structure; Cellular Automata; MACA*

## Introduction

The three-tiered structural hierarchy possessed by proteins is typically referred to as primary and tertiary structure. Protein Structure Predication from sequences of amino acid gives tremendous value to biological community. This is because the higher-level and secondary level (Abagyan et al., 1997) structures determine the function of the proteins and consequently, the insight into its function can be inferred from that.

As genome sequencing projects are increasing tremendously. The SWISS-PORT databases (Anfinsen et al., 1973) of primary protein structures are expanding tremendously. Protein Data Banks are not growing at a faster rate due to innate difficulties in finding the levels of the structures. Structure determination (Boeckmann et al., 2003) procedure experimental setups will be very expensive, time consuming, require more labor and may not applicable to all the proteins. Keeping in view of shortcomings of laboratory procedures in predicting the structure of protein major research have been dedicated to protein prediction of high level structures using computational techniques. Anfinsen did a pioneering work predicting the protein structure from amino acid sequences (Bourne et al., 2002). This is usually called as protein folding problem which is the greatest challenge in bioinformatics. This is the ability to predict the higher level structures from the amino acid sequence.

By predicting the structure of protein the topology of the chain can be described. The tree dimensional arrangement of amino acid sequences can be described by tertiary structure. They can be predicted independent of each other. Functionality of the protein can be affected by the tertiary structure, topology and the tertiary structure. Structure aids in the identification of membrane proteins, location of binding sites and identification of homologous proteins (Chandonia et al, 2001) to list a few of the benefits, and thus highlighting the importance, of knowing this level of structure This is the reason why considerable efforts have been devoted in predicting the structure only. Knowing the structure of a protein is extremely important and can also greatly enhance the accuracy of tertiary structure prediction. Furthermore, proteins can be classified according to their structural elements, specifically their alpha helix and beta sheet content.

## Related Works in Structure Prediction

The Objective of structure prediction is to identify whether the amino acid residue of protein is in helix, strand or any other shape. In 1960 as a initiative step of structure prediction the probability of respective structure element is calculated for each amino acid by taking single amino acid properties consideration (Bourne et al., 2002). This method of structure prediction is said to be first generation technique. Later this work extended by considering the local environment of amino acid said to be second generation technique. In case of particular amino acid structure prediction adjacent residues information also needed, it considers the local environment of amino acid it gives 65% structure information. So that extension work gives 60% accuracy. The third generation technique includes machine learning, knowledge about proteins, several algorithms which gives 70% accuracy. Neural networks (Chou et al.,2000) are also useful in implementing structure prediction programs like PHD, SAM-T99.

The evolution process is directed by the popular Genetic Algorithm (GA) with the underlying philosophy of survival of the fittest gene. This GA framework can be adopted to arrive at the desired CA rule structure appropriate to model a physical system. The goals of GA formulation are to enhance the understanding of the ways CA performs computations and to learn how CA may be evolved to perform a specific computational task and to understand how evolution creates complex global behavior in a locally interconnected system of simple cells.

## Cellular Automata

Cellular Automata (CA) is a simple model of a spatially extended decentralized system, made up of a number of individual components (cells). The communication among constituent cells is limited to local interaction. Each individual cell is in a specific state that changes over time depending on the states of its neighbors. From the days of Von Neumann who first proposed the model of Cellular Automata (CA), to Wolfram's recent book 'A New Kind of Science', the simple and local neighborhood structure of CA has attracted researchers from diverse disciplines.

Definition: CA is defined a four tipple <G, Z, N, F>
Where G -> Grid (Set of cells)
Z -> Set of possible cell states
N-> Set which describe cells neighborhoods
F -> Transition Function (Rules of automata)

The concept of the homogeneous structure of CA was initiated in early 1950s by J. Von Neumann. It was conceived as a general framework for modeling complex structures, capable of self-reproduction and self-repair. Subsequent developments have taken place in several phases and in different directions.

*Artificial Immune Systems*

Artificial immune systems are motivated by the theory of immunology. The biological immune system functions to protect the body against pathogens or antigens that could potentially cause harm. It works by producing antibodies that identify, bind to, and finally eliminate the pathogens. Even though the number of antigens is far larger than the number of antibodies, the biological immune system has evolved to allow it to deal with the antigens. The immune system will learn the criteria of the antigens so that in future it can react both to those antigens it has encountered before as well as to entirely new ones. In 2002, de Castro and Timmis (Irback et al., 2002) suggested that "for a system to be characterized as an artificial immune system, it has to embody at least a basic model of an immune component (e.g. cell, molecule, organ), it has to have been designed using the ideas from theoretical and/or experimental immunology.

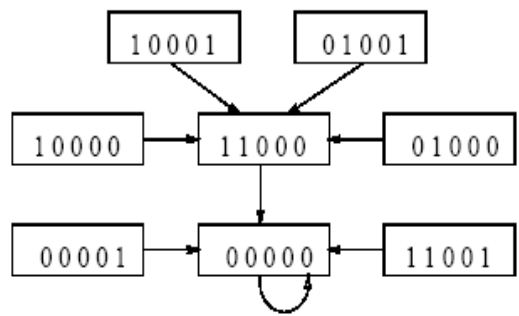

FIG. 1 EXAMPLE OF MACA WITH BASIN 0000

## Design of MACA Based Pattern Classifier with Artificial Immune System

An n-bit MACA with k-attractor basins can be viewed as a natural classifier. It classifies a given set of patterns into k number of distinct classes, each class containing the set of states in the attractor basin. To enhance the classification accuracy of the machine, most of the works have employed MACA to classify patterns into two classes (say I and II). The following example illustrates an MACA based two class pattern classifier.

*The Proposed Artificial Immune Algorithm*

The algorithm works as in Figure 2 (after each six steps we have one cell generation):

(1) Generate a set (P) of candidate solutions, composed

of the subset of memory cells (M) added to the remaining (Pr) population (P = Pr + M);

(2) Determine (Select) the n best individuals of the population (Pn), based on an affinity measure;

(3) Reproduce (Clone) these n best individuals of the population, giving rise to a temporary population of clones (C). The clone size is an increasing function of the affinity with the antigen;

(4) Submit the population of clones to a hypermutation scheme, where the hyper mutation is proportional to the affinity of the antibody with the antigen. A maturated antibody population is generated ($C^*$);

(5) Re-select the improved individuals from $C^*$ to compose the memory set M. Some members of P can be replaced by other improved members of $C^*$.

(6) Replace d antibodies by novel ones (diversity introduction). The lower affinity cells have higher probabilities of being replaced.

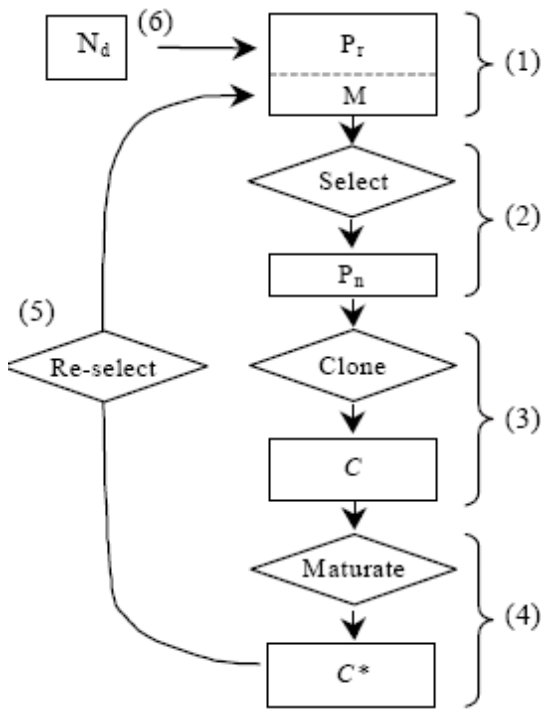

FIG. 2 PROPOSED ALGORITHM

*Hybrid Genetic-Immune System Method*

The proposed hybrid method depends on genetic algorithms and immune system. The main forces of the evolutionary process for the GA are crossover and the mutation operators. For the Clonal selection algorithm the main force of the evolutionary process is the idea of clone selection in which new clones are generated. These new clones are then mutated and the best of these clones are added to the population plus adding new generated members to the population.

The hybrid method take the main force of the evolutionary process for the two systems.

The hybrid method is described as follow:

1- Generate the initial population (candidate solutions).
2- Select the (N) best items from the population.
3- For each selected item generate a number of clones ($N_c$) and mutate each item form ($N_c$).
4- Select the best mutated item from each group ($N_c$) and add it to the population.
5- Select from the population the items on which the crossover will be applied. We select them randomly in our system but any selection method can be used.
6- After selection make a crossover and add the new items (items after crossover) to the population by replacing the low fitness items with the new ones.
7- Add to the population a group of new generated random items.
8- Repeat step 2-7 according to meeting the stopping criterion.

The steps 2-5 were repeated for a number of times before adding new group of generated random items.

**Experiments**

*Dataset Description*

The used data is recorded for a speech recognition task. The 30 samples for 9 words are collected . These words represent the digits from 1 to 9 spoken in arabic language. As a standard procedure in evaluating machine learning techniques, the dataset is split into a training set and a test set. The training set is composed of 15 x 9 utterances, and the same size is used for the test set. HMM models are trained using the above three methods. Then, the performance of each model is tested on the test dataset. Models are compared according to the average log likelihood over all utterances for each word. Moreover, HMM model is trained by using one traditional method (Baum- Welch algorithm). The results are reported in table 1.

The objective of these experiments is to determine which of four methods yields better model in terms of the maximum likelihood estimation (MLE) of training and testing data.

A special class of non-linear CA, termed as Multiple Attractor CA (SPECIAL MACA), has been proposed to develop the model. Theoretical analysis, reported in this chapter, provides an estimate of the noise accommodating capability of the proposed SPECIAL MACA based associative memory model. Characterization of the basins of attraction of the proposed model establishes the sparse network of non-linear CA (SPECIAL MACA) as a powerful pattern recognizer for memorizing unbiased patterns. It provides an efficient and cost-effective alternative to the dense network of neural net for pattern recognition. Detailed analysis of the SPECIAL MACA rule space establishes the fact that the rule subspace of the pattern recognizing/classifying CA lies at the edge of chaos. Such a CA, as projected in (Karplus et al. , 2002) , is capable of executing complex computation. The analysis and experimental results reported in the current and next chapters confirm this viewpoint. A SPECIAL MACA employing the CA rules at the edge of chaos is capable of performing complex computation associated with pattern recognition.

*Algorithm Single Point Crossover*

Input: Two randomly selected rule vectors (Parent 1 and 2).

Output: Resultant rule vectors (Offspring 1 and 2).

Step 1: Randomly generate a number 'q' in between 1 and n.

Step 2: Take the first q rules (symbols) from first rule vector (Parent 1) and the (n-q) rules of Parent 2. Form a new rule vector (Offspring 1) concatenating these rules.

Step 3: Form Offspring 2 by concatenating the first q rules of Parent 2 and the last (n-q) rules of Parent 1.

Step 4: Stop.

*Random Generation of Initial Population*

To form the initial population, it must be ensured that each solution randomly generated is a combination of an n-bit DS with 2m number of attractor basins (Classifier #1) and an m-bit DV (Classifier #2). The chromosomes are randomly synthesized according to the following steps.

1. Randomly partition n into m number of integers such that
$$n_1 + n_2 + \cdots + n_m = n.$$
2. For each $n_i$, randomly generate a valid Dependency Vector (DV).
3. Synthesize Dependency String (DS) through concatenation of m number of DVs for Classifier #1.
4. Randomly synthesize an m-bit Dependency Vector (DV) for Classifier #2.
5. Synthesize a chromosome through concatenation of Classifier #1 and Classifier #2.

**Experimental Step**

- Select the target CA protein (amino acid sequence) T, whose structure is to be predicted.
- Perform a AIS-MACA search, using the primary amino acid sequence Tp of the target CA protein T. The objective is being to locate a set of CA proteins, S = {S1, S2…} of similar sequence
- Select from S the primary structure Bp of a base CA protein, with a significant match to the target CA protein. A AIS-MACA (Pradipta Maji et al. , 2003) search produces a measure of similarity between each CA protein in S and the target CA protein T. Therefore, Bp can be chosen as the CA protein with the highest such value
- Obtain the base CA protein's structure, Bs, from the PDB
- Using Bp, create an input sequences Ib (corresponding to the base CA protein) by replacing each amino acid in the primary structure with its hydrophobia city value. The output sequences Ob is created by replacing the structural elements in Bs with the values, 200, 600, 800 for helix C, strand and coil respectively
- Solve the system identification problem, by performing CA de convolution with the output sequences Ob and the input sequence Ib to obtain the CA response, or the sought after running the algorithm.
- Transform the amino acid sequence of Tp into a discrete time sequences It, and convolve with F; thereby producing the predicted structure (Ot = It*F) of the target CA protein
- The result of this calculation Ot is a vector of numerical values. For values between 0 and 200, a helix C is predicted, and between 600 and 800, a strand is predicted by CA. All other values will be predicted as a coil by MACA. This produces mapping for the required target structure Ts of the target CA protein T

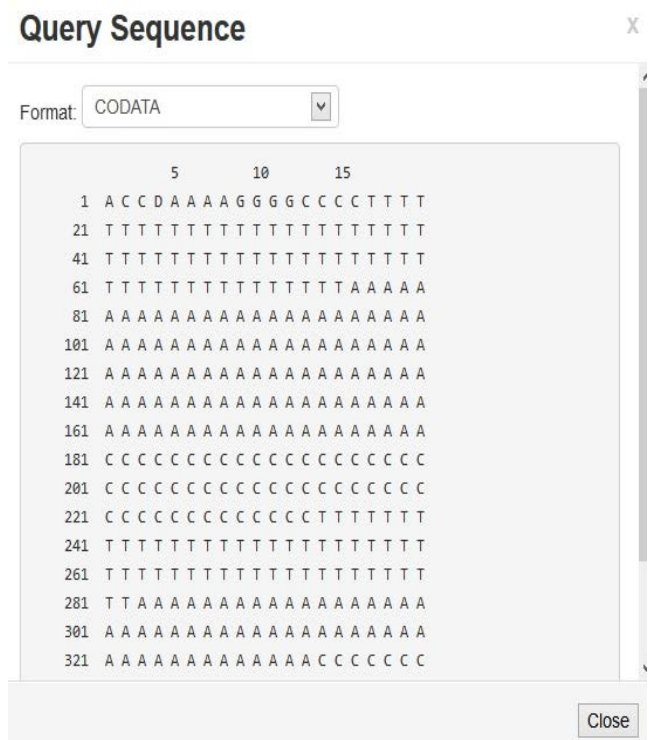

FIG. 3 AMINO ACID SEQUENCE

## Experimental Results

In the experiments conducted, the base proteins are assigned the values 400,800,1000 for helix C, strand and coil respectively. We have found an structure numbering scheme that is build on Boolean characters of CA which predicts the coils, stands and helices separately .The MACA based prediction procedure as described in the previous section is then executed, and each occurrence of each sequences in the resulting output, is predicted . The query sequence analyzer was designed and identification of the green terminals of the protein is simulated in the figure 4. The analysis of the sequence and the place of joining of the proteins are also pointed out in the figure 5. Experimental results Figure 7, 8 which include the similarity and accuracy graph with each of the components are separately plotted.

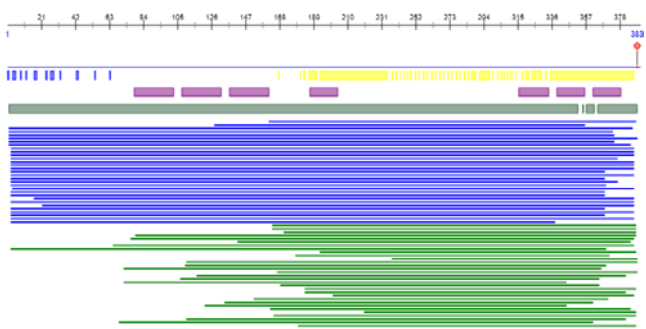

FIG. 4 PROTEIN STRUCTURE PREDICTION INTERFACE WITH GREEN AS POSITIVE

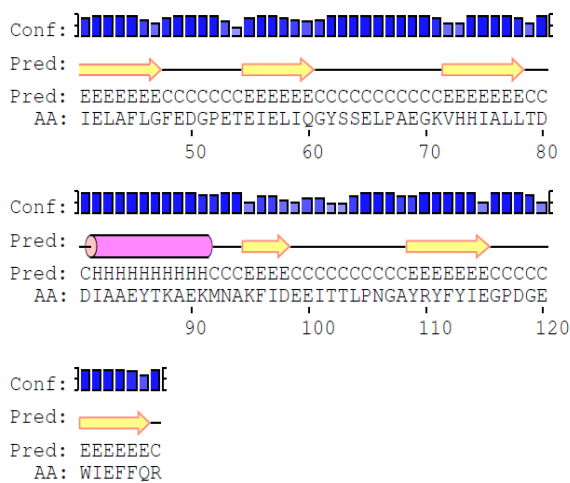

FIG. 5 PROTEIN STRUCTURE PREDICTION ANALYSIS

| Target: 1PFC | Prediction Accuracy | Target: 1PP2 | Prediction Accuracy | Target: 1QL8 | Prediction Accuracy |
|---|---|---|---|---|---|
| Exp 1 | 65% | Exp 5 | 85% | Exp 9 | 85% |
| Exp 2 | 65% | Exp 6 | 90% | Exp 10 | 90% |
| Exp 3 | 69% | Exp 7 | 83% | Exp 11 | 82% |
| Exp 4 | 71% | Exp 8 | 87% | Exp 12 | 91% |

FIG. 6 PREDICTION ACCURACY

| Prediction Method | Prediction Accuracy for 1PFC | Prediction Accuracy for 1PP2 | Prediction Accuracy for 1QL8 |
|---|---|---|---|
| DSP | 92% | 70% | 96% |
| PHD | 70% | 68% | 84% |
| SAM-T99 | 68% | 77% | 87% |
| SS Pro | 70% | 73% | 81% |
| AIS-MACA | 90% | 85% | 97% |
| AIS-AIS-MACA | 92% | 83% | 96% |

FIG. 7 PREDICTION ACCURACY FOR AIS--AIS-MACA

## Conclusion

Existing structure-prediction methods can predict the structure of protein with 70% accuracy. Artificial Immune System was employed not only to increase the accuracy of prediction but also for identifying proteins in overlatped seqauences also, thus strenghtning the system.To provide a more thorough analysis of the viability of our proposed technique more experiments will be conducted .Our results indicate that such a level of accuracy is attainable, and can be potentially surpassed with our method. AIS-MACA provides the best overall accuracy that ranges between 80% and 89.8% depending on the dataset.

### REFERENCES

Abagyan, R., Batalov S., Cardozo,T., Totrov, M., Webber, J., Zhou, Y. 1997. Homology Modeling With Internal Coordinate Mechanics: Deformation Zone Mapping and Improvements of Models via Conformational Search.

PROTEINS: Structure, Function and Genetics. 1:29-37

Alexandrov, N., Solovyev, V., 1996. Effect of secondary structure prediction on protein fold recognition and database search. Genome Informatics 7, 119-127

Anfinsen, C. B., 1973. Principles that govern the folding of protein chains. Science. 181, 223-230.

Baldi, P., Brunak, S., Frasconi, P., Pollastri, G., Soda, G., 2000. Bidirectional Dynamics for Protein Secondary Structure Prediction. Sequence Learning: Paradigms, Algorithms and Applications. Springer, 80-104

Boeckmann B., Bairoch A., Apweiler R., Blatter M.-C., Estreicher A., Gasteiger E., Martin M.J., Michoud K., O'Donovan C., Phan I., Pilbout S., Schneider M., 2003. *The SWISS-PROT protein knowledgebase and its supplement TrEMBL in 2003* Nucleic Acids Res. 31:365-370.

Bonneau, R., Tsai, J., Ruczinski, I., Chivian, D., Rohl, C., Strauss, C., Baker, D. 2001. Rosetta in CASP4: Progress in Ab Initio Protein Structure Prediction. PROTEINS: Structure, Function and Genetics. 5:119-126

Bourne, Philip E., Weissig, Helge, 2003. Structural Bioinformatics. John Wiley & Sons.

Brandon C., Tooze J., 1999. Introduction to Protein Structure. Garland Publishing.

Chandonia, J., Karplus M., 1999. New Methods for Accurate Prediction of Protein Secondary Structure. PROTEINS: Structure, Function and Genetics, 35, 293-306

Chou, P., Fasman G., 1978. Prediction of the secondary structure of proteins from their amino acid sequence. Advanced Enzymology, 47, 45-148

Dunbrack, R.. 1999. Comparative Modeling of CASP3 Targets Using PSI-BLAST and SCWRL. PROTEINS: Structure, Function and Genetics 3:81-87

Debasis Mitra, Michael Smith, "Digital Signal Processing in Protein Secondary Structure Prediction" Innovations in Applied Artificial Intelligence Lecture Notes in Computer Science Volume 3029, 2004.

Eric E. Snyder ,Gary D. Stormo, " Identification of Protein Coding Regions In Genomic DNA".ICCS Transactions 2002.

E E Snyder and G D Stormo,"Identification of coding regions in genomic DNA sequences: an application of dynamic programming and neural networks " Nucleic Acids Res. 1993 February 11; 21(3): 607–613.

Hirakawa, H., Kuhara, S., 1997. Prediction of Hydrophobic Cores of Proteins Using Wavelet Analysis. Genome Informatics, 8, 61-70

Irback, A., Sandelin, E., 2000 On Hydrophobicity Correlations in Protein Chains. Biophysical Journal, 79, 2252-2258

Irback, A., Peterson, C., Potthast, F., 1996. Evidence for nonrandom hydrophobicity structures in protein chains. Proc. Natl. Acad. Sci., 93, September, 9533-9538

Jadwiga Bienkowsk, Rick Lathrop ," THREADING ALGORITHMS".

Kabsch W., Sander C., 1983 Dictionary of Protein Secondary Structure: Pattern Recognition of Hydrogen-Bonded and Geometrical Features. Biopolymers, 3, 2577-2638

Karplus. K., Barrett, C., Hughey, R. 1998. Hidden Markov Models for Detecting Remote Protein Homologies. Bioinformatics, vol. 14, no. 10, 846-856

P.Kiran Sree & Dr Inampudi Ramesh Babu et al," PSMACA: An Automated Protein Structure Prediction using MACA (Multiple Attractor Cellular Automata)", accepted for publication in Journal of Bioinformatics and Intelligent Control (JBIC) in Volume 2 Number 3, (American Scientific Publications, USA)

P.Kiran Sree, I .Ramesh Babu ,"Identification of Protein Coding Regions in Genomic DNA Using Unsupervised FMACA Based Pattern Classifier" in International Journal of Computer Science & Network Security with ISSN: 1738-7906 Volume Number: Vol.8, No.1,2008.

P. Flocchini, F. Geurts, A. Mingarelli, andN. Santoro (2000),"Convergence and Aperiodicity in Fuzzy Cellular Automata: Revisiting Rule 90,"Physica D.

P. Maji and P. P. Chaudhuri (2004),"FMACA: A Fuzzy Cellular Automata Based Pattern Classifier," Proceedings of 9th International Conference on Database Systems , Korea, pp. 494–505, 2004.

P. Kiran Sree, G.V.S. Raju, I. Ramesh Babu ,S. Viswanadha Raju" Improving Quality of Clustering using Cellular Automata for Information retrieval" in International Journal of Computer Science 4 (2), 167-171, 2008. ISSN 1549-3636. ( Science Publications-USA)

P.Kiran Sree, I. Ramesh Babu"Face Detection from still and Video Images using Unsupervised Cellular Automata with K means clustering algorithm" ICGST International

Journal on Graphics, Vision and Image Processing(GVIP),ISSN: 1687-398X, Volume 8, Issue II, 2008, (1-7).

P. Maji and P. P. Chaudhuri (2004),"FMACA: A Fuzzy Cellular Automata Based Pattern Classifier," Proceedings of 9th International Conference on Database Systems , Korea, 2004, pp. 494–505.

P. Maji and P. P. Chaudhuri, "Fuzzy Cellular Automata For Modeling Pattern Classifier," *IEICE,* (2004).

R. Lippmann, An introduction to computing with neural nets, IEEE ASSP Mag. 4(22) (2004), pp.121-129.

Skolnick, J., Kolinski, A., 2001. Computational Studies of Protein Folding. Computing in Science and Engineering. September/October, Vol. 3, No. 5, 40-49

Thiele, R., Zimmer, R., Lengauer, T. Protein Threading by Recursive Dynamic Programming. Journal of Molecular Biology 290, 757-779

Veljkovic V, Cosic I, Dimitrijevic B, Lalovic D., 1985. Is It Possible To Analyze DNA and Protein Sequences by the Methods of Digital Signal Processing, IEEE Transactions on Biomedical Engineering, Vol. BME-32, No. 5, 337-341, 1985